\documentclass[runningheads,a4paper]{llncs}
\usepackage{amsmath}
\usepackage{mwe}    % loads »blindtext« and »graphicx«
\usepackage{tikz}
\usepackage{subfig}
\usepackage{palatino}
\usepackage{array}
\usepackage{parskip}
\newcolumntype{P}[1]{>{\centering\arraybackslash}p{#1}}
\newcolumntype{M}[1]{>{\centering\arraybackslash}m{#1}}
\usepackage{url}
\urldef{\mailsa}\path|{alfred.hofmann, ursula.barth, ingrid.haas, frank.holzwarth,|
\urldef{\mailsb}\path|anna.kramer, leonie.kunz, christine.reiss, nicole.sator,|
\urldef{\mailsc}\path|erika.siebert-cole, peter.strasser, lncs}@springer.com|    
\newcommand{\keywords}[1]{\par\addvspace\baselineskip
\noindent\keywordname\enspace\ignorespaces#1}

\begin{document}
\mainmatter  % start of an individual contribution

\title{An Introduction to Convolutional Neural Networks}

\titlerunning{Introduction to Convolutional Neural Networks}

\author{Keiron O'Shea\inst{1} \and Ryan Nash\inst{2}}
%2
\authorrunning{Keiron O'Shea et al.} % abbreviated author list (for running head)
\tocauthor{}
\institute{
  Department of Computer Science, Aberystwyth University, Ceredigion, SY23 3DB\\
  \email{keo7@aber.ac.uk}
  \and
  School of Computing and Communications, Lancaster University, Lancashire, LA1 4YW\\
  \email{nashrd@live.lancs.ac.uk}
}

\maketitle

\begin{abstract}
The field of machine learning has taken a dramatic twist in recent times, with the rise of the Artificial Neural Network (ANN). These biologically inspired computational models are able to far exceed the performance of previous forms of artificial intelligence in common machine learning tasks. One of the most impressive forms of ANN architecture is that of the Convolutional Neural Network (CNN). CNNs are primarily used to solve difficult image-driven pattern recognition tasks and with their precise yet simple architecture, offers a simplified method of getting started with ANNs.

This document provides a brief introduction to CNNs, discussing recently published papers and newly formed techniques in developing these brilliantly fantastic image recognition models. This introduction assumes you are familiar with the fundamentals of ANNs and machine learning.

\end{abstract}

\section{Introduction}

\textbf{Artificial Neural Networks} (ANNs) are computational processing systems of which are heavily inspired by way biological nervous systems (such as the human brain) operate. ANNs are mainly comprised of a high number of interconnected computational nodes (referred to as neurons), of which work entwine in a distributed fashion to collectively learn from the input in order to optimise its final output.

\begin{figure}[h]
\centering
\includegraphics[width=0.65\textwidth]{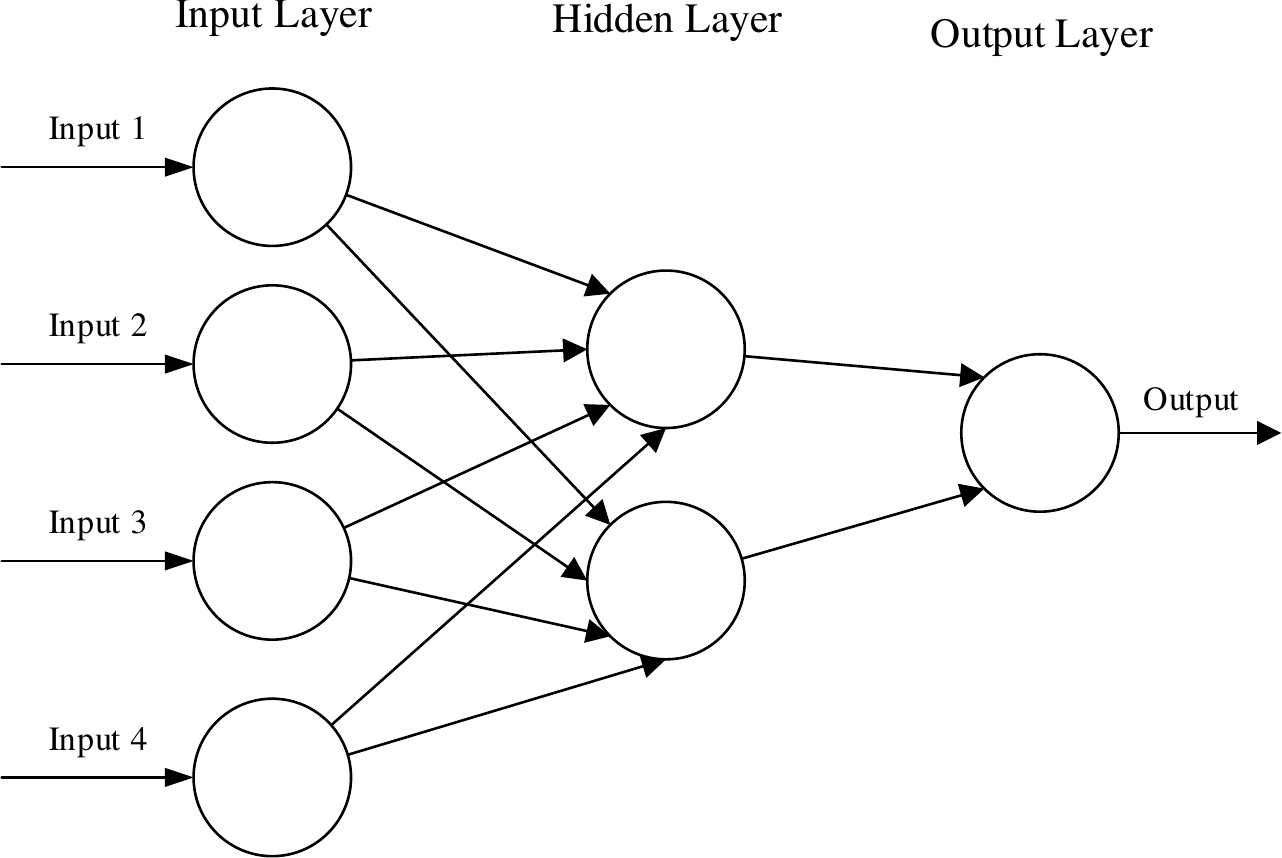}
\caption{A simple three layered feedforward neural network (FNN), comprised of a input layer, a hidden layer and an output layer. This structure is the basis of a number of common ANN architectures, included but not limited to Feedforward Neural Networks (FNN), Restricted Boltzmann Machines (RBMs) and Recurrent Neural Networks (RNNs).}
\label{fig:threelayerneuralnetwork}
\end{figure}

The basic structure of a ANN can be modelled as shown in Figure \ref{fig:threelayerneuralnetwork}. We would load the input, usually in the form of a multidimensional vector to the input layer of which will distribute it to the hidden layers. The hidden layers will then make decisions from the previous layer and weigh up how a stochastic change within itself detriments or improves the final output, and this is referred to as the process of learning. Having multiple hidden layers stacked upon each-other is commonly called deep learning.

The two key learning paradigms in image processing tasks are supervised and unsupervised learning. \textbf{Supervised learning} is learning through pre-labelled inputs, which act as targets. For each training example there will be a set of input values (vectors) and one or more associated designated output values. The goal of this form of training is to reduce the models overall classification error, through correct calculation of the output value of training example by training.

\textbf{Unsupervised learning} differs in that the training set does not include any labels. Success is usually determined by whether the network is able to reduce or increase an associated cost function. However, it is important to note that most image-focused pattern-recognition tasks usually depend on classification using supervised learning.

\textbf{Convolutional Neural Networks} (CNNs) are analogous to traditional ANNs in that they are comprised of neurons that self-optimise through learning. Each neuron will still receive an input and perform a operation (such as a scalar product followed by a non-linear function) - the basis of countless ANNs. From the input raw image vectors to the final output of the class score, the entire of the network will still express a single perceptive score function (the weight). The last layer will contain loss functions associated with the classes, and all of the regular tips and tricks developed for traditional ANNs still apply.

The only notable difference between CNNs and traditional ANNs is that CNNs are primarily used in the field of pattern recognition within images. This allows us to encode image-specific features into the architecture, making the network more suited for image-focused tasks - whilst further reducing the parameters required to set up the model.

One of the largest limitations of traditional forms of ANN is that they tend to struggle with the computational complexity required to compute image data. Common machine learning benchmarking datasets such as the MNIST database of handwritten digits are suitable for most forms of ANN, due to its relatively small image dimensionality of just $28\times28$. With this dataset a single neuron in the first hidden layer will contain $784$ weights ($28\times28\times1$ where $1$ bare in mind that MNIST is normalised to just black and white values), which is manageable for most forms of ANN. 

If you consider a more substantial coloured image input of $64\times64$, the number of weights on just a single neuron of the first layer increases  substantially to $12,288$. Also take into account that to deal with this scale of input, the network will also need to be a lot larger than one used to classify colour-normalised MNIST digits, then you will understand the drawbacks of using such models.

\subsection{Overfitting}

But why does it matter? Surely we could just increase the number of hidden layers in our network, and perhaps increase the number of neurons within them? The simple answer to this question is no. This is down to two reasons, one being the simple problem of not having unlimited computational power and time to train these huge ANNs.

The second reason is stopping or reducing the effects of overfitting. \textbf{Overfitting} is basically when a network is unable to learn effectively due to a number of reasons. It is an important concept of most, if not all machine learning algorithms and it is important that every precaution is taken as to reduce its effects. If our models were to exhibit signs of overfitting then we may see a reduced ability to pinpoint generalised features for not only our training dataset, but also our test and prediction sets.

This is the main reason behind reducing the complexity of our ANNs. The less parameters required to train, the less likely the network will overfit - and of course, improve the predictive performance of the model.

\section{CNN architecture}

As noted earlier, CNNs primarily focus on the basis that the input will be comprised of images. This focuses the architecture to be set up in way to best suit the need for dealing with the specific type of data.

One of the key differences is that the neurons that the layers within the CNN are comprised of neurons organised into three dimensions, the spatial dimensionality of the input (\textbf{height} and the \textbf{width}) and the \textbf{depth}. The depth does not refer to the total number of layers within the ANN, but the third dimension of a activation volume. Unlike standard ANNS, the neurons within any given layer will only connect to a small region of the layer preceding it.

In practice this would mean that for the example given earlier, the input 'volume' will have a dimensionality of $64\times64\times3$ (height, width and depth), leading to a final output layer comprised of a dimensionality of $1\times1\times n$ (where $n$ represents the possible number of classes) as we would have condensed the full input dimensionality into a smaller volume of class scores filed across the depth dimension.

\subsection{Overall architecture}

\begin{figure}[h]
\centering
\includegraphics[width=0.65\textwidth]{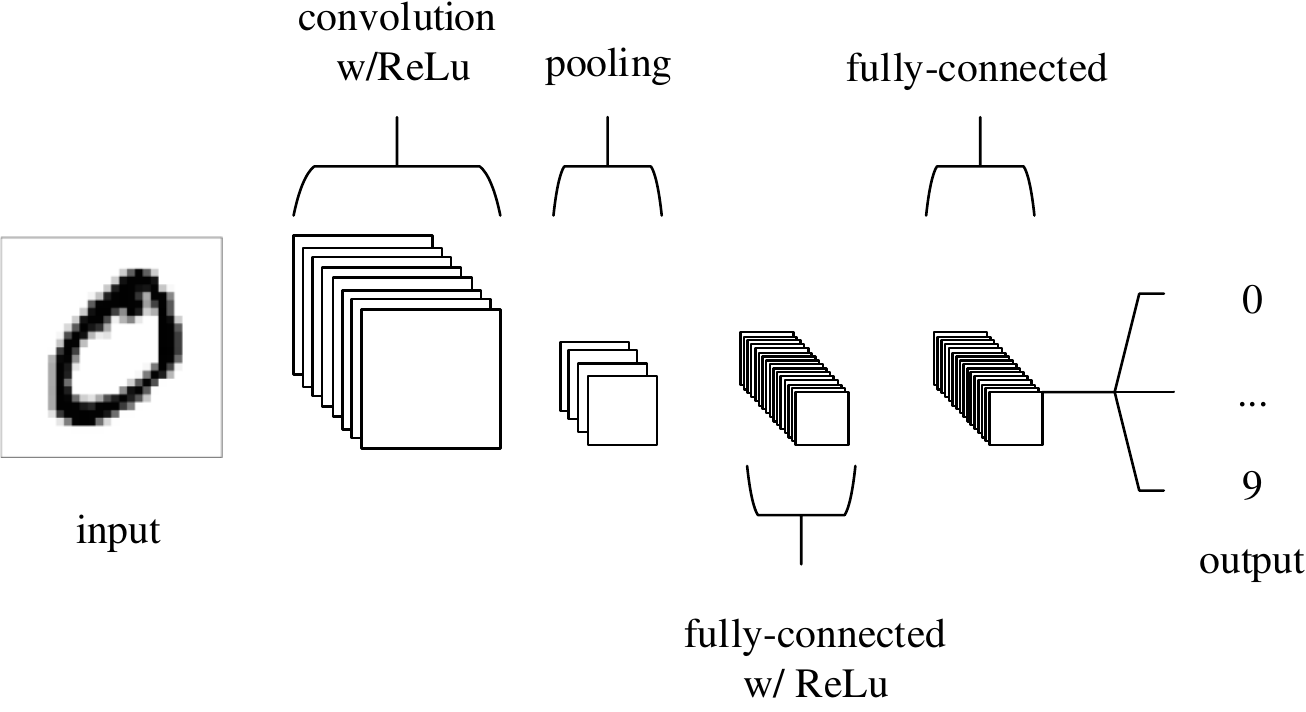}
\caption{An simple CNN architecture, comprised of just five layers}
\label{fig:simplecnn}
\end{figure}

CNNs are comprised of three types of layers. These are convolutional layers, pooling layers and \textbf{fully-connected layers}. When these layers are stacked, a CNN architecture has been formed. A simplified CNN architecture for MNIST classification is illustrated in Figure \ref{fig:simplecnn}.

The basic functionality of the example CNN above can be broken down into four key areas.

\begin{enumerate}
\item As found in other forms of ANN, the \textbf{input layer} will hold the pixel values of the image. 
\item The \textbf{convolutional layer} will determine the output of neurons of which are connected to local regions of the input through the calculation of the scalar product between their weights and the region connected to the input volume. The \textbf{rectified linear unit} (commonly shortened to ReLu) aims to apply an 'elementwise' activation function such as sigmoid to the output of the activation produced by the previous layer. 
\item The \textbf{pooling layer} will then simply perform downsampling along the spatial dimensionality of the given input, further reducing the number of parameters within that activation. 
\item The \textbf{fully-connected layers} will then perform the same duties found in standard ANNs and attempt to produce class scores from the activations, to be used for classification. It is also suggested that ReLu may be used between these layers, as to improve performance.
\end{enumerate}

Through this simple method of transformation, CNNs are able to transform the original input layer by layer using convolutional and downsampling techniques to produce class scores for classification and regression purposes.
 
\begin{figure}[h]
\centering
\includegraphics[width=\textwidth]{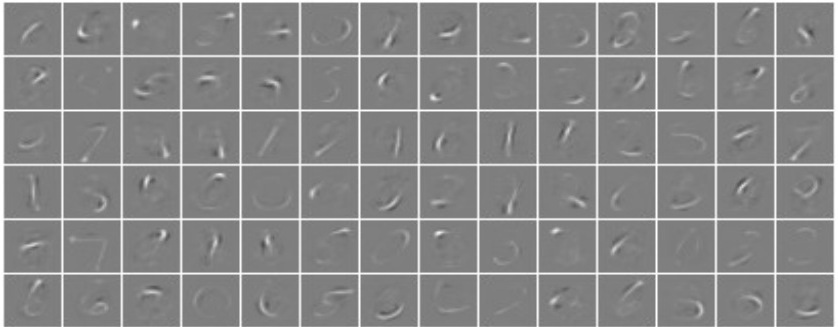}
\caption{Activations taken from the first convolutional layer of a simplistic deep CNN, after training on the MNIST database of handwritten digits. If you look carefully, you can see that the network has successfully picked up on characteristics unique to specific numeric digits.}
\label{fig:activationmap}
\end{figure}

However, it is important to note that simply understanding the overall architecture of a CNN architecture will not suffice. The creation and optimisation of these models can take quite some time, and can be quite confusing. We will now explore in detail the individual layers, detailing their hyperparameters and connectivities.

\subsection{Convolutional layer}

As the name implies, the convolutional layer plays a vital role in how CNNs operate. The layers parameters focus around the use of learnable \textbf{kernels}.

These kernels are usually small in spatial dimensionality, but spreads along the entirety of the depth of the input. When the data hits a convolutional layer, the layer convolves each filter across the spatial dimensionality of the input to produce a 2D activation map. These activation maps can be visualised, as seen in Figure \ref{fig:activationmap}.

\begin{figure}
\centering
\includegraphics[width=0.8\textwidth]{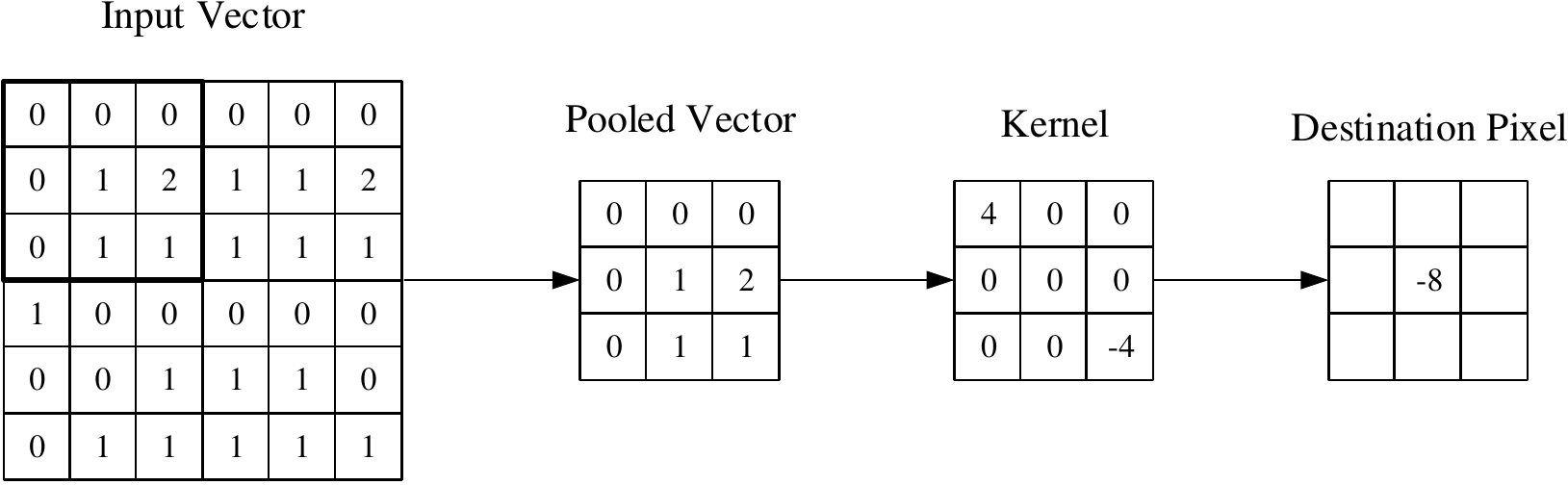}
\caption{A visual representation of a convolutional layer. The centre element of the kernel is placed over the input vector, of which is then calculated and replaced with a weighted sum of itself and any nearby pixels.}
\label{fig:visual}
\end{figure}

As we glide through the input, the scalar product is calculated for each value in that kernel. (Figure \ref{fig:visual}) From this the network will learn kernels that 'fire' when they see a specific feature at a given spatial position of the input. These are commonly known as \textbf{activations}.

Every kernel will have a corresponding activation map, of which will be stacked along the depth dimension to form the full output volume from the convolutional layer.

As we alluded to earlier, training ANNs on inputs such as images results in models of which are too big to train effectively. This comes down to the fully-connected manner of standard ANN neurons, so to mitigate against this every neuron in a convolutional layer is only connected to small region of the input volume. The dimensionality of this region is commonly referred to as the \textbf{receptive field size} of the neuron. The magnitude of the connectivity through the depth is nearly always equal to the depth of the input.

For example, if the input to the network is an image of size $64\times64\times3$ (a RGB-coloured image with a dimensionality of $64\times64$) and we set the receptive field size as $6\times6$, we would have a total of $108$ weights on each neuron within the convolutional layer. ($6\times6\times3$ where $3$ is the magnitude of connectivity across the depth of the volume) To put this into perspective, a standard neuron seen in other forms of ANN would contain $12,288$ weights each.

Convolutional layers are also able to significantly reduce the complexity of the model through the optimisation of its output. These are optimised through three hyperparameters,  the \textbf{depth}, the \textbf{stride} and setting \textbf{zero-padding}.

The \textbf{depth} of the output volume produced by the convolutional layers can be manually set through the number of neurons within the layer to a the same region of the input. This can be seen with other forms of ANNs, where the all of the neurons in the hidden layer are directly connected to every single neuron beforehand. Reducing this hyperparameter can significantly minimise the total number of neurons of the network, but it can also significantly reduce the pattern recognition capabilities of the model.

We are also able to define the \textbf{stride} in which we set the depth around the spatial dimensionality of the input in order to place the receptive field. For example if we were to set a stride as 1, then we would have a heavily overlapped receptive field producing extremely large activations. Alternatively, setting the stride to a greater number will reduce the amount of overlapping and produce an output of lower spatial dimensions.

\textbf{Zero-padding} is the simple process of padding the border of the input, and is an effective method to give further control as to the dimensionality of the output volumes.

It is important to understand that through using these techniques, we will alter the spatial dimensionality of the convolutional layers output. To calculate this, you can make use of the following formula:

\begin{equation*}
(V - R) + 2Z \over S + 1
\end{equation*}

Where $V$ represents the input volume size ($ \mathrm{height} \times \mathrm{width} \times \mathrm{depth}$), $R$ represents the receptive field size, $Z$ is the amount of zero padding set and $S$ referring to the stride. If the calculated result from this equation is not equal to a whole integer then the stride has been incorrectly set, as the neurons will be unable to fit neatly across the given input.

Despite our best efforts so far we will still find that our models are still enormous if we use an image input of any \textit{real} dimensionality. However, methods have been developed as to greatly curtail the overall number of parameters within the convolutional layer.

\textbf{Parameter sharing} works on the assumption that if one region feature is useful to compute at a set spatial region, then it is likely to be useful in another region. If we constrain each individual activation map within the output volume to the same weights and bias, then we will see a massive reduction in the number of parameters being produced by the convolutional layer.

As a result of this as the backpropagation stage occurs, each neuron in the output will represent the overall gradient of which can be totalled across the depth - thus only updating a single set of weights, as opposed to every single one.

\subsection{Pooling layer}

Pooling layers aim to gradually reduce the dimensionality of the representation, and thus further reduce the number of parameters and the computational complexity of the model.

The pooling layer operates over each activation map in the input, and scales its dimensionality using the ``MAX'' function. In most CNNs, these come in the form of \textbf{max-pooling layers} with kernels of a dimensionality of $2\times2$ applied with a stride of $2$ along the spatial dimensions of the input. This scales the activation map down to 25\% of the original size - whilst maintaining the depth volume to its standard size.

Due to the destructive nature of the pooling layer, there are only two generally observed methods of max-pooling. Usually, the stride and filters of the pooling layers are both set to $2\times2$, which will allow the layer to extend through the entirety of the spatial dimensionality of the input. Furthermore \textbf{overlapping pooling} may be utilised, where the stride is set to $2$ with a kernel size set to $3$. Due to the destructive nature of pooling, having a kernel size above $3$ will usually greatly decrease the performance of the model.

It is also important to understand that beyond max-pooling, CNN architectures may contain general-pooling. \textbf{General pooling} layers are comprised of pooling neurons that are able to perform a multitude of common operations including L1/L2-normalisation, and average pooling. However, this tutorial will primarily focus on the use of max-pooling. 

\subsection{Fully-connected layer}

The fully-connected layer contains neurons of which are directly connected to the neurons in the two adjacent layers, without being connected to any layers within them. This is analogous to way that neurons are arranged in traditional forms of ANN. (Figure \ref{fig:threelayerneuralnetwork})

\section{Recipes}

Despite the relatively small number of layers required to form a CNN, there is no set way of formulating a CNN architecture. That being said, it would be idiotic to simply throw a few of layers together and expect it to work. Through reading of related literature it is obvious that much like other forms of ANNs, CNNs tend to follow a common architecture. This common architecture is illustrated in Figure \ref{fig:simplecnn}, where convolutional layers are stacked, followed by pooling layers in a repeated manner before feeding forward to fully-connected layers.

\begin{figure}[h]
\centering
\includegraphics[width=\textwidth]{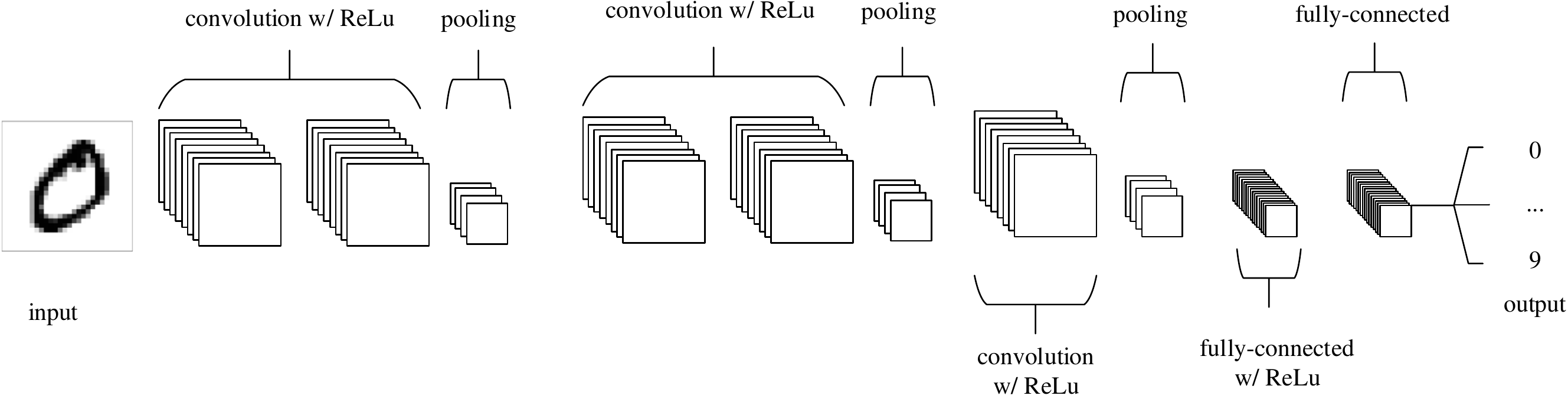}
\caption{A common form of CNN architecture in which convolutional layers are stacked between ReLus continuously before being passed through the pooling layer, before going between one or many fully connected ReLus.}
\label{fig:deepcnn}

\end{figure}

Another common CNN architecture is to stack two convolutional layers before each pooling layer, as illustrated in Figure \ref{fig:deepcnn}. This is strongly recommended as stacking multiple convolutional layers allows for more complex features of the input vector to be selected.

It is also advised to split large convolutional layers up into many smaller sized convolutional layers. This is to reduce the amount of computational complexity within a given convolutional layer. For example, if you were to stack three convolutional layers on top of each other with a receptive field of $3\times3$. Each neuron of the first convolutional layer will have a $3\times3$ view of the input vector. A neuron on the second convolutional layer will then have a $5\times5$ view of the input vector. A neuron on the third convolutional layer will then have a $7\times7$ view of the input vector. As these stacks feature non-linearities which in turn allows us to express stronger features of the input with fewer parameters. However, it is important to understand that this does come with a distinct memory allocation problem - especially when making use of the backpropagation algorithm.

The input layer should be recursively divisible by two. Common numbers include $32\times32$, $64\times64$, $96\times96$, $128\times128$ and $224\times224$.

Whilst using small filters, set stride to one and make use of zero-padding as to ensure that the convolutional layers do not reconfigure any of the dimensionality of the input. The amount of zero-padding to be used should be calculated by taking one away from the receptive field size and dividing by two.activation

CNNs are extremely powerful machine learning algorithms, however they can be horrendously resource-heavy. An example of this problem could be in filtering a large image (anything over $128\times128$ could be considered large), so if the input is $227\times227$ (as seen with ImageNet) and we're filtering with 64 kernels each with a zero padding of then the result will be three activation vectors of size $227\times227\times64$ - which calculates to roughly 10 million activations - or an enormous 70 megabytes of memory per image. In this case you have two options. Firstly, you can reduce the spatial dimensionality of the input images by resizing the raw images to something a little less heavy. Alternatively, you can go against everything we stated earlier in this document and opt for larger filter sizes with a larger stride (2, as opposed to 1).

In addition to the few rules-of-thumb outlined above, it is also important to acknowledge a few 'tricks' about generalised ANN training techniques. The authors suggest a read of Geoffrey Hinton's excellent ``Practical Guide to Training Restricted Boltzmann Machines''.

\section{Conclusion}

Convolutional Neural Networks differ to other forms of Artifical Neural Network in that instead of focusing on the entirety of the problem domain, knowledge about the specific type of input is exploited. This in turn allows for a much simpler network architecture to be set up.

This paper has outlined the basic concepts of Convolutional Neural Networks, explaining the layers required to build one and detailing how best to structure the network in most image analysis tasks.

Research in the field of image analysis using neural networks has somewhat slowed in recent times. This is partly due to the incorrect belief surrounding the level of complexity and knowledge required to begin modelling these superbly powerful machine learning algorithms. The authors hope that this paper has in some way reduced this confusion, and made the field more accessible to beginners.

\section*{Acknowledgements}

The authors would like to thank Dr. Chuan Lu and Nicholas Dimonaco for useful discussion and suggestions.

\end{document}